\documentclass[a4paper,twoside]{article}

\usepackage{epsfig}
\usepackage{subcaption}
\usepackage{calc}
\usepackage{amssymb}
\usepackage{amstext}
\usepackage{amsmath}
\usepackage{amsthm}
\usepackage{multicol}
\usepackage{mathptmx} 
\usepackage{apalike}
\usepackage{algorithm2e}
\usepackage[bottom]{footmisc}

\usepackage{SCITEPRESS} 

\DeclareMathOperator*{\argmin}{arg\,min}

\usepackage{float}
\begin{document}

\title{Yolo-Key-6D : Single Stage Monocular 6D Pose Estimation with Keypoint Enhancements}

\author{\authorname{Kemal Alperen Çetiner\sup{1,2}\orcidAuthor{0009-0003-0021-317X}, Hazım Kemal Ekenel\sup{2,3}\orcidAuthor{0000-0003-3697-8548}}
\affiliation{\sup{1}ASELSAN, Aerospace Technologies, Türkiye}
\affiliation{\sup{2}Istanbul Technical University, Department of Computer Engineering, Türkiye}
\affiliation{\sup{3}New York University Abu Dhabi, Division of Engineering, UAE}
\email{{\{cetiner23, ekenel\}}@itu.edu.tr}
}


\abstract{Estimating the 6D pose of objects from a single RGB image is a critical task for robotics and extended reality applications. However, state-of-the-art multi stage methods often suffer from high latency, making them unsuitable for real time use. In this paper, we present Yolo-Key-6D, a novel single stage, end-to-end framework for monocular 6D pose estimation designed for both speed and accuracy. Our approach enhances a YOLO based architecture by integrating an auxiliary head that regresses the 2D projections of an object's 3D bounding box corners. This keypoint detection task significantly improves the network's understanding of 3D geometry. For stable end-to-end training, we directly regress rotation using a continuous 9D representation projected to $SO(3)$ via singular value decomposition. On the LINEMOD and LINEMOD-Occluded benchmarks, YOLO-Key-6D achieves competitive accuracy scores of 96.24\% and 69.41\%, respectively, with the ADD(-S) 0.1d metric, while proving itself to operate in real time. Our results demonstrate that a carefully designed single stage method can provide a practical and effective balance of performance and efficiency for real world deployment.}

\maketitle 
\normalsize 
\setcounter{footnote}{0}

\section{\uppercase{Introduction}}
\label{sec:introduction}

The problem of estimating the 3D rigid pose of an object from an RGB image entails calculating the camera's pose relative to the object, expressed as a combination of a 3D rotation and a 3D translation \cite{survey1}. Solutions to this problem have wide ranging applications in robotics \cite{robot1,robot2}, from object grasping to robot localization. Another field where this problem is critical is extended reality (XR), particularly for applications involving spatial manipulation \cite{xr}.

6 DoF (Degree of Freedom) pose estimation remains a highly challenging problem due to a combination of visual and geometric complexities. Key difficulties include heavy occlusion, object symmetry, and strong variations in lighting conditions, as well as the need to handle diverse objects with different shapes and appearances. Additional challenges arise from textureless or reflective surfaces, scale ambiguity when depth cues are limited, and cluttered backgrounds that may contain visually similar distractors. A prevailing approach is to leverage deep neural networks to overcome these challenges. State of the art performance is predominantly achieved via multi stage pose estimation strategies. The conventional approach involves an initial establishment of 2D-3D correspondences, followed by 6D pose determination using a RANSAC based Perspective-n-Point (PnP) algorithm \cite{cpdn,vivier}. An alternative strategy is the render and compare methodology, which leverages synthetic rendering for pose verification or refinement \cite{megapose,mrcnet}. Meanwhile, some methods further incorporate a pixel wise, RANSAC based voting scheme to bolster the robustness of keypoint localization \cite{pvnet}.

\begin{figure}[t] 
\centering 
\includegraphics[width=\columnwidth]{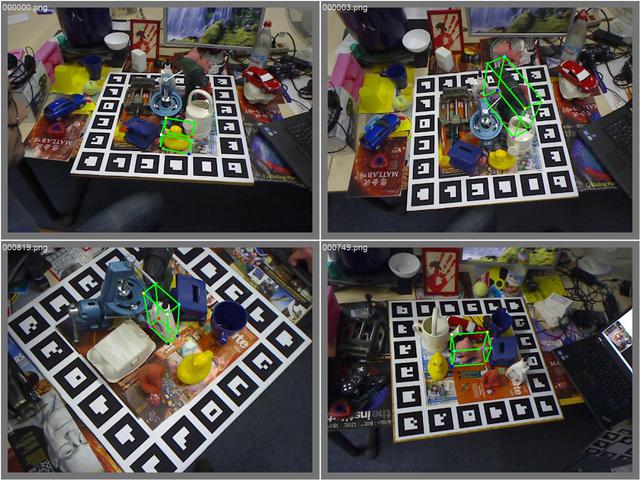} \caption{3D bounding boxes calculated from the predicted rotation matrices and translation vectors for various objects in the test dataset.} 
\label{fig:preds} 
\end{figure}

However, these multi stage approaches are difficult to employ in robotics and XR applications for several inherent reasons. First, they require extensive inference times due to their multi stage nature. They often have a stage where keypoints are extracted or a separate object detector running before pose estimation \cite{cosypose,mrcnet}, which can greatly reduce runtime speed. This introduces several problems; for example, low speed pose estimation in an XR system can cause motion sickness as a consequence of high motion to photon latency \cite{m2p}. Another issue with the multi stage approach is that it lacks end-to-end trainability. This means that methods may require intermediate steps like RANSAC or further refinements, which in turn increases inference and training times. Moreover, some multi stage methods, especially those based on keypoint extraction, have an inference time that increases linearly with the number of objects in a scene. Therefore, although multi stage approaches can provide very accurate pose estimations, their limited inference speed can be a bottleneck for fast moving devices.

To cope with the aforementioned problems of multi stage approaches, we propose a single stage pose estimation network whose prediction samples are provided in Figure \ref{fig:preds}. Even without complex intermediate steps, it can achieve accuracy comparable to multi stage methods with a significantly lower inference time. The core of our method is based on the YOLO model, augmented with heads for depth, rotation, and 3D bounding box regression as an auxiliary task. Regressing 3D bounding boxes enhances the network's understanding of the 3D world. We validate this hypothesis in our experiments, showing that it boosts predicted pose accuracy. Furthermore, we employ the $\mathbb{R}^9+SVD$ (Singular Value Decomposition)  representation for rotations, as it is better suited for following the gradient flow on the $SO(3)$ manifold \cite{learning3d}. The proposed solution is compared with the other state of the art methods for the problem of 6 DoF pose estimation of seen objects from RGB images with given 3D models. 

To summarize, the primary contributions of this paper are as follows:
\begin{itemize}
    \item We devise a single stage network that regresses the object pose while using 3D bounding box detection as an auxiliary task.
    \item We employ the $\mathbb{R}^9 + SVD$ representation for rotations, in contrast to other works that use quaternions and Euler angles.
    \item We propose a loss function consisting of distinct components corresponding to each of the network's outputs.
\end{itemize}

The remainder of this paper is organized as follows. Section~\ref{sec:Related} provides a review of related work in 6D pose estimation. While Section~\ref{sec:method} details our proposed methodology, including the rigid body parameterization, data augmentation strategies, model architecture, and loss function. Later, Section~\ref{sec:exps} presents our experimental setup, evaluation results on benchmark datasets, and an ablation study. Finally, Section~\ref{sec:conc} concludes the paper with a summary of our findings.

\section{Related Works}
\label{sec:Related}
The estimation of 6 DoF object pose from a single RGB image is a long standing problem. While traditional methods relied on local feature correspondences derived from classical keypoint extractions, the advent of deep learning has catalyzed a new generation of more robust and accurate approaches. Our review focuses on these recent deep learning based methods, which can be broadly categorized into direct estimation, render and compare refinement, and correspondence based techniques.

\subsection{Direct Pose Estimation}
\label{Direct}
Direct estimation methods formulate 6 DoF pose estimation as a regression problem, training a neural network to map input image pixels directly to a 6D pose vector ($SE(3)$ transformation). This end-to-end approach was first proposed in seminal works like PoseNet \cite{posenet} and later adapted for specific objects in PoseCNN \cite{posecnn}, demonstrating the feasibility of using CNNs for this task. The primary limitation of these early direct regression methods, however, is the highly non-linear and often ambiguous nature of the pose space, which can cause networks to converge to poor local minima.

To create a more robust learning target, many subsequent methods moved away from pure regression and framed the task as pose classification, discretizing the continuous pose space into a finite set of bins. For instance, SSD-6D \cite{ssd6d} treats rotational regression as a classification task to improve training stability. A common and effective extension is a hybrid architecture that combines a coarse pose classifier with a local regression network to refine the final estimate. 

Our work contributes to this area by proposing a direct regression approach based on anchorless YOLO framework that is end-to-end trainable. It is designed to achieve the stability needed for the pose estimation task while maintaining the  simplicity of regressing the pose directly.

\subsection{Render \& Compare Methods}
Another prominent strategy involves iteratively refining a pose estimate by minimizing the discrepancy between the input image and a synthetic image rendered using traditional graphics pipelines. This render and compare loop is a powerful technique for achieving high precision results \cite{mrcnet}. DeepIM \cite{deepim} was a foundational work in this area, demonstrating an iterative process where a network predicts the pose update by comparing the observed and rendered images. This concept was generalized to multiview settings by CosyPose \cite{cosypose}, which globally refines poses by minimizing multi-view reprojection error.

Recent advancements in this category have focused on improving the robustness and applicability of the refinement process. For example, MegaPose \cite{megapose} introduced a refiner that can be applied to novel objects by taking synthetic views of the objects' CAD models as input. Furthermore, FocalPose \cite{focalpose} extended this paradigm by jointly estimating the camera's focal length along with the object's pose, addressing a critical variable in real world scenarios. A defining characteristic of these methods is their reliance on an initial coarse pose and their iterative nature. 

For the sake of completeness one recent development in this category is the use of differentiable renderers like Pythorch3D \cite{torch3d}, which enables gradient based optimization directly through the rendering process for fine grained pose refinement. One of the impactful on this subcategory is iNerf inverts the direction of a trained Nerf to obtain 6D pose estimates \cite{inerf}.   
\subsection{Correspondence Based Methods}
The most prevalent paradigm for 6 DoF pose estimation involves a two stage process: first, a network predicts 2D-3D correspondences between image pixels and points on the object's 3D model; second, a PnP algorithm computes the pose from these matches \cite{Hartley}.

Early methods in this category focused on detecting a sparse set of semantic keypoints. BB8 \cite{bb8}, for example, regressed the 2D projections of a 3D bounding box's corners. To further enhance robustness, more recent techniques predict dense correspondences, establishing a link between every visible object pixel and its coordinate on the 3D model's surface. DPODv2 \cite{DPODv2} exemplifies this trend, relying on dense correspondence estimation and a multi view refinement stage.

Another recent development in this area is the emergence of zero shot and few shot learning approaches. Methods like POPE (Promptable Object Pose Estimation)  \cite{pope} enable 6 DoF pose estimation for any object with only a single reference view, leveraging the power of foundation models. 

The primary challenge for all correspondence based methods is their reliance on a PnP solver, which is a non-differentiable geometric algorithm. This breaks the end-to-end training pipeline, preventing the correspondence network from receiving direct gradient feedback based on the final pose error and potentially leading to suboptimal performance. While recent work like EPro-PnP \cite{epro} has proposed techniques to create a differentiable PnP layer by proposing it as a probabilistic structure, this adds architectural complexity to the solution. On the other hand, our method uses a direct regression which enables it to be fully differentiable. Therefore, it is able to utilize the benefits of the end-to-end training while being architecturally simple.

\section{Methodology}
\label{sec:method}
Given that most use cases of 6 DoF pose estimation revolve around robotics and XR, a method is needed that can meet requirements such as low computational cost, high speed, and high precision. Our work presents an end-to-end framework that adheres to these requirements. The framework estimates the pose $P=[R\vert t]$ of a known object from a single RGB image, where $R$ is the rotation matrix and $t$ is the translation vector, given the object's 3D CAD model, by employing a series of strategies.

As the first step of our method, we embrace a rigorous parameterization and representation of rigid body transformations to ensure stability. Following this, a focused analysis is performed on various 6D data augmentation techniques, acknowledging the difficulty of generating valid, non-constrained pose variations. Furthermore, the proposed model has an architecture that can leverage these parameterizations and augmentations to generate robust, real time estimations. To this end, we introduce several model heads, each specializing in a separate aspect of the problem. Finally, we propose a loss function that is a combination of individual loss functions suitable for the different model heads.

\subsection{Rigid Body Parameterization}
Any rigid body transformation lies on the $SE(3)$ manifold, which has properties distinct from standard Euclidean space. Therefore, a proper parameterization is required to effectively learn and represent the problem's underlying geometry.

\subsubsection{Rotation Parameterization}There are multiple ways to represent 3D rotations, including Euler angles, quaternions, and matrices. However, a common issue with these representations is ambiguity, where different parameters can describe the same rotation. Examples of this include gimbal lock in Euler angles and the double cover problem in quaternions \cite{rotdisc}.

To avoid issues like gimbal lock during training, representations such as unit quaternions or log quaternions are often used \cite{deepim}. The main drawback of quaternions is that they create a double cover over the rotation group $SO(3)$, meaning two distinct quaternions, $q$ and $-q$, represent the same orientation. This can introduce discontinuities when treated naively in Euclidean space \cite{rotdisc}. To address this, \cite{6dposerep} employs a 6D representation based on Gram-Schmidt Orthonormalization (GSO), where two 3D vectors are regressed and their cross product yields the third one. However, this method is known to bias the resulting rotation matrix towards the first two predicted vectors, which can lower the precision of the final orientation \cite{learning3d}.

Consequently, we employ a method that finds the closest valid rotation on the $SO(3)$ manifold via singular value decomposition \cite{learning3d}. The formulation is based on the intuition that a regressed 9D vector, when reshaped into a 3x3 matrix, is not guaranteed to be a valid rotation matrix, i.e. an element of $SO(3)$. We can assume this regressed matrix has been perturbed from a pure rotation. The problem then becomes: "What is the valid rotation closest to the regressed matrix?". To solve this, we employ the Orthogonal Procrustes solution \cite{procrustes}, defined in Equation \ref{eq:procrustes}.

\begin{equation} \label{eq:procrustes}
\begin{aligned}
    &R^* = \argmin_{R} ||R - M||_{F} \quad s.t. \quad R^{T}R = I \\
    &M = U\Sigma V^T \\
    &R^* = UV^T
\end{aligned}
\end{equation}
Here, $R^* \in SO(3)$ is the closest valid rotation matrix to $M$, and $M \in M_{3,3}(\mathbb{R})$ is the regressed, potentially non-orthogonal matrix. 

\subsubsection{Translation Parameterization}
Directly regressing the translation vector $t = [t_x, t_y, t_z] \in \mathbb{R}^3$ is challenging because objects can appear anywhere within the image, which creates a large and unbounded solution space for the regression parameters \cite{posecnn}. Therefore, inspired by the work of \cite{posecnn} and \cite{gdr}, we determine the object's position by separately estimating its 2D projection $(o_x, o_y)$ on the image plane and its distance $t_z$ from the camera. Afterwards, given the camera's intrinsic matrix $K$, the full 3D translation vector $t$ can be recovered using backprojection, as shown in Equation \ref{eq:backprojection}.

\begin{equation} 
\label{eq:backprojection}
\begin{aligned}
    &t = t_z K^{-1} [o_x, o_y, 1]^T
\end{aligned}
\end{equation}
However, since estimating depth from a single RGB image is an ill posed task, we constrain the problem by reformulating the depth prediction as shown in Equation \ref{eq:depth}.
\begin{equation}
\label{eq:depth}
\begin{aligned}
&t_z = dist_{min} + \sigma (dist_{max} - dist_{min})
\end{aligned}
\end{equation}
Here, $[dist_{min}, dist_{max}]$ represents the known range of possible object distances from the camera for a given dataset. Thus, instead of regressing the absolute depth $t_z$, our model only needs to predict a normalized scale factor $\sigma \in [0,1]$.

\subsection{Augmentations}
Given the nature of the problem, augmentations can be grouped into two categories: one that relates to the image domain and the other pertaining to 3D world. 
\subsubsection{Image Domain Augmentations} 
\begin{figure}[t] 
\centering 
\includegraphics[width=0.8\columnwidth]{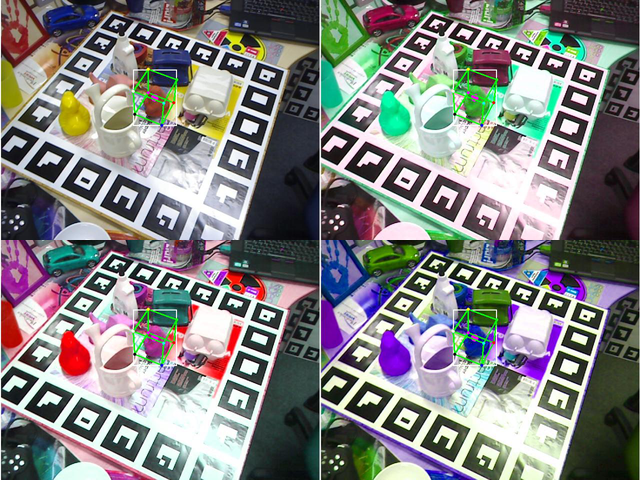} \caption{Augmented data sample with changed HSV values. Top left image is the original and remaining ones are augmented samples.} 
\label{fig:hsv} 
\end{figure}

Using the HSV color space improves model robustness by allowing for independent changes to the gains of the brightness (Value) and color (Hue, Saturation) channels, as shown in Figure~\ref{fig:hsv}. 
Mathematically, this is achieved by applying random gains to each channel of an image after its conversion to the HSV color space. 
Applying a gain to the Value channel effectively simulates different lighting conditions according to the Equation~\ref{eq:val}.
\begin{equation}\label{eq:val}
    V_{\text{new}} = V_{\text{orig}} \times (1 + r_v \times \text{hsv\_v})
\end{equation}
Here, $V_{\text{orig}}$ is the original value, $r_v$ is a random number in the range $[-1, 1]$, and $\text{hsv\_v}$ is the maximum augmentation factor. 
Similarly, adjusting gains on the Hue and Saturation channels creates realistic color variations given in Equation \ref{eq:hs}.
\begin{equation}\label{eq:hs}
\begin{aligned}
    H_{\text{new}} &= (H_{\text{orig}} + r_h \times \text{hsv\_h}) \pmod{1} \\
    S_{\text{new}} &= S_{\text{orig}} \times (1 + r_s \times \text{hsv\_s})
\end{aligned}
\end{equation}
The new Hue ($H_{\text{new}}$) is wrapped using a modulo operation, while the new Saturation ($S_{\text{new}}$) and Value ($V_{\text{new}}$) are clipped to the range of $[0, 1]$. Training a model on these augmented images teaches it to recognize an object's core features, thereby improving its performance in real world scenarios~\cite{vivier}.
\begin{figure}[t]
  \centering
   \includegraphics[width=0.8\columnwidth]{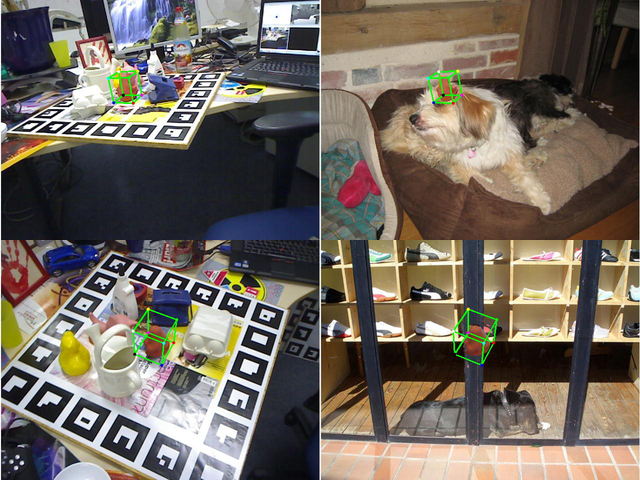}
  \caption{Object of interest is cutout and the background is replaced with an image from VOC 2012 dataset. Left images are originals and ones on right are augmented images.}
  \label{fig:bg}
 \end{figure}

Moreover, to prevent the model from becoming biased towards the specific environments in the training datasets, we replace the background of the object images. We utilize images from the VOC 2012 dataset \cite{pascal-voc-2012} for this background replacement, which is illustrated in Figure \ref{fig:bg}. 

\subsubsection{3D Augmentations}
\begin{figure}[t]
  \centering
   \includegraphics[width=0.8\columnwidth]{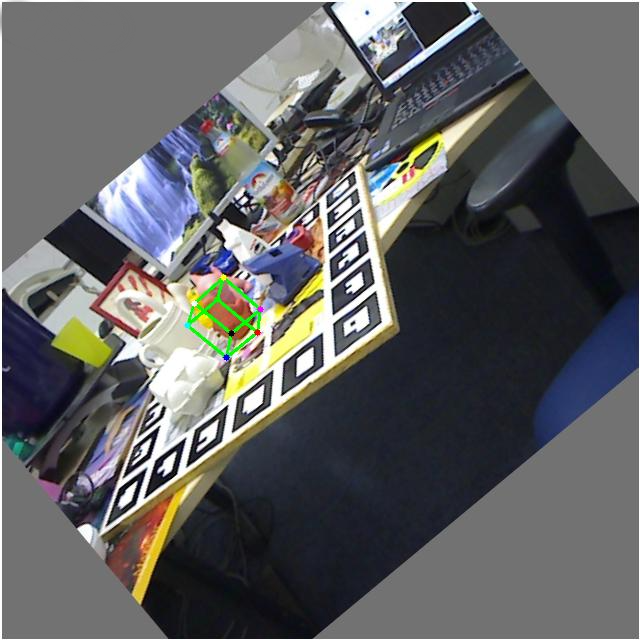}
  \caption{Image is rotated around the principal axis.}
  \label{fig:rotaug}
 \end{figure}

To derive a 3D augmentation strategy, we seek 3D transformations that produce a consistent and predictable effect in the 2D image plane; these are known as equivariant transformations. We explore equivariant transformations since a desired property of any augmentation is that it must preserve the validity of the ground truth labels to be useful in the learning process. We can examine the projection function $P$ to identify such transformations. The projection of a 3D point $X_{object}$ to a 2D image point $x_{image}$ is defined by Equation \ref{eq:projection}.
\begin{equation}\label{eq:projection}
    sx_{image} = K(R^{object}_{cam}X_{object}+t) \equiv P(X_{object})
\end{equation}
Here, $s$ represents the depth of the point in the camera's coordinate system and $K$ is the intrinsic matrix of the camera.
A specific transformation that satisfies this equivariance property is the rotation of the object around the camera's principal axis, as illustrated in Figure \ref{fig:rotaug}. The principal axis is the camera's optical axis, which conventionally corresponds to the Z-axis in the camera's local coordinate system.

A rotation around the principal (Z) axis by an angle $\theta$ is represented by the matrix:
\begin{equation}\label{eq:rz}
    R_z(\theta) = \begin{pmatrix} \cos\theta & -\sin\theta & 0 \\ \sin\theta & \cos\theta & 0 \\ 0 & 0 & 1 \end{pmatrix}
\end{equation}
When we apply this to a point $X_{cam} = [X, Y, Z]^T$, we get the transformed point $X'_{cam}$ given in Equation~\ref{eq:rot_rz}.
\begin{equation}\label{eq:rot_rz}
    X'_{cam} = R_z(\theta) X_{cam} = \begin{pmatrix} X\cos\theta - Y\sin\theta \\ X\sin\theta + Y\cos\theta \\ Z \end{pmatrix}
\end{equation}
The key observation here is that the third component, the depth $Z$, remains unchanged ($Z' = Z$). Since the scale factor $s$ in the projection equation is precisely this depth value, the depth of every point remains constant under this specific 3D rotation. This property allows us to define a consistent 2D transformation in the image plane via the homography $H$ given in Equation~\ref{eq:homo}.
\begin{equation}\label{eq:homo}
    x'_{image} = H x_{image} \quad \text{where} \quad H = K R_z(\theta) K^{-1}
\end{equation}
 
In the common case where the camera has square pixels and a centered principal point, this homography simplifies to a pure 2D rotation of the image around its center. This direct correspondence between a 3D object rotation and a 2D image rotation makes it an ideal equivariant transformation for data augmentation. In contrast, other transformations, such as general affine transforms, would require objects to be planar or very far from the camera for the homography in Equation \ref{eq:homo} to hold \cite{Hartley}.

\subsection{Model}
We introduce a YOLOv11 based architecture tailored for estimating the 6D pose of an object. In contrast to two stage methods that first detect objects and then estimate their 6D pose, our model performs both tasks in a single stage. This integrated approach enables our model to be a complete end-to-end trainable solution.

YOLOv11 was selected as our base detector due to its state-of-the-art balance of speed and accuracy, providing a powerful foundation for real time applications like XR. Our goal is to augment this architecture for 6D pose estimation while introducing minimal computational overhead. The network leverages the Extended Efficient Layer Aggregation Networks (E-ELAN) based backbone and a programmable, gradient path aware neck structure inherent to the model \cite{yolov11}.

\begin{figure*}[t]
    \centering
    \includegraphics[width=\textwidth]{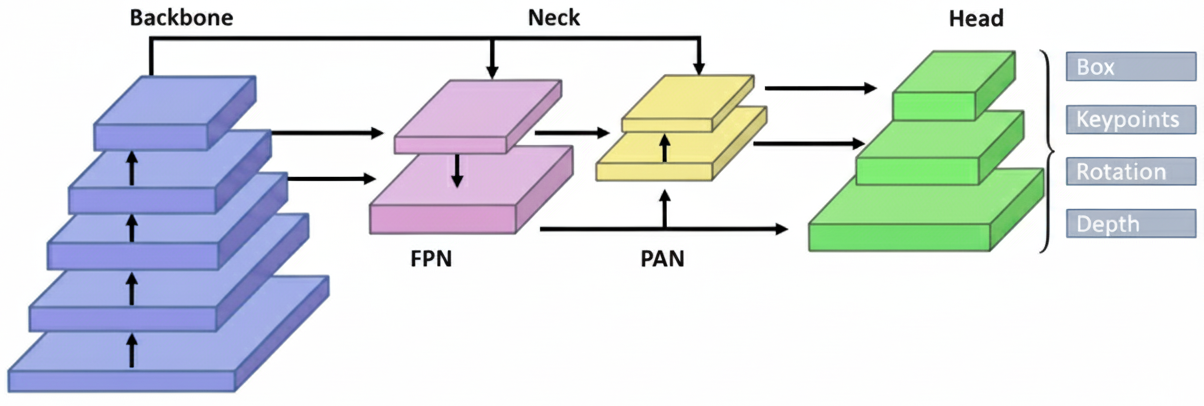}
    \caption{Yolov11 combines E-ELAN backbone using group convolutions with FPN and PAN neck layers to create aggregate features. Our model produces Rotation, Depth and Keypoint vectors in addition to usual detection head in order to perform the 6 DoF Pose Estimation task.}
    \label{fig:model}
\end{figure*}

To enable pose estimation, we integrate dedicated heads for rotation and 3D localization alongside the standard detection heads of the network, as illustrated in Figure \ref{fig:model}.

The rotation head is trained to predict a continuous 9D representation of the object's orientation, as described in Equation \ref{eq:procrustes}. This representation is subsequently decoded into a final 3x3 rotation matrix using a post processing step. Notably, our rotation head is a single shot predictor and does not rely on iterative refinement modules.

The keypoint head is trained to predict the 2D projections of the 3D bounding box corners, as well as the object's center. Additionally, occlusion information is predicted for each keypoint to enable the model to better handle occluded objects.

\subsection{Loss Function}
The total loss function is composed of distinct components, each tailored to a different task and output head of the model. The main components are the rotation loss, translation loss, keypoint loss, and 2D bounding box loss.

\subsubsection{Rotation Loss $(\mathcal{L}_{R})$}
The geodesic distance on the $SO(3)$ manifold measures the shortest "rotation angle" between two 3D orientations. Therefore, our rotation loss, defined in Equation \ref{eq:geodesic}, calculates the geodesic distance between the predicted rotation ($R_{\text{pred}}$) and the ground truth rotation ($R_{\text{gt}}$). This provides a natural way to measure the angular error between the two rotations \cite{rotdisc}.

\begin{equation} \label{eq:geodesic}
\begin{aligned}
d(R_{\text{pred}}, R_{\text{gt}}) = \arccos\left(\frac{\text{trace}(R_{\text{pred}}^T R_{\text{gt}}) - 1}{2}\right)
\end{aligned}
\end{equation}

\subsubsection{Translation Loss $(\mathcal{L}_{t})$}
As described in the parameterization section, we simplify the depth estimation task by having the model predict a normalized scale factor instead of an absolute distance. For this regression task, we employ the smooth $L1$ loss, defined in Equation \ref{eq:smoothL1}, to penalize errors in the predicted scale.

\begin{equation} \label{eq:smoothL1}
\text{SmoothL1}(y, \hat{y}) = 
\begin{cases} 
0.5 \cdot (y - \hat{y})^2 / \beta & \text{if } |y - \hat{y}| < \beta \\
|y - \hat{y}| - 0.5 \cdot \beta & \text{otherwise} 
\end{cases}
\end{equation}
Where, $y$ is the true scale, $\hat{y}$ is the predicted scale and $\beta$ is set to 1.

\subsubsection{Keypoint Loss $(\mathcal{L}_{kp})$}
We formulate the 3D bounding box prediction as a keypoint detection task, where the keypoints correspond to the projected corners of the box and the object's center. Our loss function is inspired by the Object Keypoint Similarity (OKS) metric used in the COCO challenge \cite{coco}.
\begin{equation} \label{eq:kploss}
\begin{aligned}
    log(\mathcal{L}_{kp}) &= \frac{1}{B} \sum_{j=1}^{B} \left[ \left( \frac{N_{\text{kp}}}{\sum_{k=1}^{N_{kp}} v_{kj}} \right) \sum_{i=1}^{N_{kp}} v_{ij} \cdot L_{ij} \right] \\
    B &: Batch\, Size \\
    N_{kp} &: Number \, of \, Keypoints \,(9) \\ 
    v &: visibility\\
    L &: l_2 \: distance
\end{aligned}
\end{equation}
As shown in Equation \ref{eq:kploss}, the loss is a weighted L2 distance between predicted and ground truth keypoints. A visibility mask ($v$) is applied to ensure that occluded or unannotated keypoints do not contribute to the loss.

\subsubsection{Bounding Box Loss $(\mathcal{L}_{bb})$}
The 2D bounding box loss merges two concepts: Complete IoU (CIoU) loss and Distribution Focal Loss (DFL) \cite{yolov11}. The CIoU component, shown in Equation \ref{eq:bboxloss}, evaluates the geometric alignment by considering overlap, center point distance, and aspect ratio. The DFL component complements this by treating the box coordinates as a probability distribution, which helps the model learn the boundaries more precisely.

\begin{equation} \label{eq:bboxloss}
\begin{aligned}
\mathcal{L}_{bb} = 1 - IoU + \frac{\rho^2(b, b_{gt})}{c^2} + \alpha v
\end{aligned}
\end{equation}
The formula calculates the loss between a predicted box  $b$ and a ground truth box  $b_{gt}$, where $\rho^2(b, b_{gt})$ is the squared distance between their centers, $c$ is the diagonal of their smallest enclosing box, and $\alpha$ is a parameter that weights $v$, the measure of aspect ratio consistency.
Finally, the total loss function is a weighted sum of these individual components, as defined in Equation \ref{eq:finalloss}. The $\lambda$ terms are hyperparameters that control the contribution of each loss component.

\begin{equation} \label{eq:finalloss}
\begin{aligned}
\mathcal{L}_{total} = \lambda_{R}\mathcal{L}_{R}+\lambda_{t}\mathcal{L}_{t}+\lambda_{kp}\mathcal{L}_{kp}+\lambda_{bb}\mathcal{L}_{bb}
\end{aligned}
\end{equation}

\section{Experiments}
\label{sec:exps}
We perform our experiments on the LINEMOD \cite{hinterstoisser2012model} and LINEMOD-Occluded \cite{brachmann2014learning} benchmark datasets. We evaluate our results using the ADD(-S) 0.1d metric \cite{hinterstoisser2012model}, where a pose estimate is considered correct if its ADD or ADD-S score is less than 10\% of the object's diameter. For symmetric objects, the ADD-S score is calculated using Equation \ref{eq:adds}, where $M$ is the set of the object's 3D model points and $m$ is the total number of points.
\begin{equation}\label{eq:adds}
    ADD\text{-}S = \frac{1}{m} \sum_{x_1 \in M} \min_{x_2 \in M} \|(Rx_1 + t) - (\hat{R}x_2 + \hat{t})\|
\end{equation}
We trained and evaluated our model for each object in the benchmark datasets individually, using the provided test sets for evaluation. We consider the average score across all objects to be the primary indicator of the model's overall performance.

\begin{table}[ht]
\caption{Proposed Yolo-Key-6D's performance using 10\% ADD and ADD-S metric for LINEMOD (LM) and LINEMOD-Occluded (LM-O) datasets.}
\label{tab:dataset_results_avg}
\centering
\begin{tabular}{l | c | c | c}
\hline
\textbf{Object} & \textbf{Diameter (m)} & \textbf{LM} & \textbf{LM-O} \\
\hline
Ape & 0.10 & 99.6 & 72.6 \\
Benchvise & 0.25 & 91.0 & - \\
Bowl & 0.17 & 95.3 & - \\
Cam & 0.17 & 96.3 & - \\
Can & 0.20 & 98.3 & 68.4 \\
Cat & 0.15 & 97.0 & 71.7 \\
Cup & 0.12 & 98.0 & - \\
Driller & 0.26 & 92.8 & 70.3 \\
Duck & 0.11 & 92.4 & 67.4 \\
Eggbox & 0.16 & 99.2 & 65.6 \\
Glue & 0.18 & 93.8 & 71.0 \\
Holepuncher & 0.15 & 97.6 & 68.3 \\
Iron & 0.28 & 95.6 & - \\
Lamp & 0.28 & 98.3 & - \\
Phone & 0.21 & 98.4 & - \\
\hline
\textbf{Average} & \textbf{0.20} & \textbf{96.24} & \textbf{69.41} \\
\hline
\end{tabular}
\end{table}

We also performed an experiment to determine if our method can provide real time pose estimation. Although results are device specific, we achieved around 63 FPS during inference on a machine with a 12 GB RTX 4080, as shown in Table \ref{tab:timing}.  This result demonstrates that our model can support real time operations in XR systems equipped with graphics cards.  

\begin{table}[h]
\centering
\caption{Processing time for different operations.}
\label{tab:timing}
\begin{tabular}{l|c}
\hline
\textbf{Operation} & \textbf{Time (ms)} \\
\hline
Preprocess  & 0.8               \\
Prediction  & 13.1              \\
Postprocess & 2.1               \\
\hline
\textbf{Total} & \textbf{16.0}  \\
\hline
\end{tabular}
\end{table}

Some of the methods listed in Table \ref{tab:lm_comparison_sorted} include refinement stages, which prevent them from processing input in real time. Moreover, models like SO-Pose \cite{sopose} require a separate object detector to crop the object region, increasing the system's computational load and latency. In contrast, while our model's overall performance drops on the occluded dataset, it is still able to outperform other methods. This may signify that adding 3D bounding boxes as visibility cues is beneficial for overcoming the negative consequences of occlusions.

\begin{table}[h]
\centering
\caption{Comparison of recent methods with ADD(-S) 0.1d metric on LM and LM-O datasets.}
\label{tab:lm_comparison_sorted}
\begin{tabular}{l|c|c}
\hline
\textbf{Methods} & \textbf{LM} & \textbf{LM-O} \\
\hline
RNNPose \cite{rnnpose}     & 97.37  & 60.65  \\
Implicit Pose \cite{s24175721} & 97.25  & -      \\
\textbf{Yolo-Key-6D} & \textbf{96.24} & \textbf{69.41} \\
RePose \cite{repose}      & 96.1   & 51.6   \\
SO-Pose \cite{sopose}     & 94.0   & 62.3   \\
GDR-Net \cite{gdr}     & 93.7   & 62.2   \\
Hai et. al \cite{10376790}     & 92.2   & 65.4   \\
Hybridpose \cite{HybridPose}    & 91.3   & 47.5   \\
CDPN \cite{cpdn}        & 89.86  & -      \\
DeepIM \cite{deepim}      & 88.6   & 55.5   \\
PVNet \cite{pvnet}       & 86.27  & 40.77  \\
SSD-6D \cite{ssd6d}      & 79.0   & -      \\
\hline
\end{tabular}
\end{table}

\begin{table}[h]
\centering
\caption{Estimated Cost and Parameters for different models with a 640x640 RGB Input.}
\label{tab:pose_gflops_params}
\begin{tabular}{l|c|c}
\hline
\textbf{Method} & \textbf{GFLOP} & \textbf{Par. (M)} \\
\hline
\textbf{Yolo-Key-6D}          & \textbf{7.3}                  & \textbf{2.85}           \\
Hai et. al \cite{10376790}    & 15+            & 12+  \\
\cite{10550665}               & 31.2           & 11.6  \\
RePose \cite{repose}          & 35+            & 22+  \\
CDPN \cite{cpdn}              & 40+            & 25+    \\
PVNet \cite{pvnet}            & 57             & 11.2 \\
RNNPose \cite{rnnpose}        & 85             & 30   \\
SO-Pose \cite{sopose}         & 120+           & 15+  \\
\hline
\end{tabular}
\end{table}

\subsection{Ablation Study}
The results of our ablation study in Table \ref{tab:dataset_results_kp} demonstrate the critical role of auxiliary keypoint detection in robust 6D pose estimation. The model variant incorporating a keypoint detection head achieved a mean accuracy of 96.24\%, whereas its removal caused a collapse in performance to 76.73\%. This significant degradation highlights the ill-posed nature of inferring 3D properties from a single 2D image. The keypoint detection task mitigates this inherent ambiguity by imposing strong geometric constraints on the model.

\begin{table}[ht]
\caption{Comparison of LINEMOD with Keypoint head and no keypoint head Results for Yolo-Key-6D with ADD(-S) 0.1d.}
\label{tab:dataset_results_kp}
\centering
\begin{tabular}{l|c|c}
\hline
\textbf{Object} & \textbf{LM-kp} & \textbf{LM-no-kp} \\
\hline
Ape & 99.6 & 95.6 \\
Benchvise & 91.0 & 83.3 \\
Bowl & 95.3 & 91.9 \\
Cam & 96.3 & 89.2 \\
Can & 98.3 & 79.6 \\
Cat & 97.0 & 78.0 \\
Cup & 98.0 & 76.6 \\
Driller & 92.8 & 64.9 \\
Duck & 92.4 & 57.7 \\
Eggbox & 99.2 & 59.7 \\
Glue & 93.8 & 65.6 \\
Holepuncher & 97.6 & 74.8 \\
Iron & 95.6 & 75.3 \\
Lamp & 98.3 & 77.2 \\
Phone & 98.4 & 81.5 \\
\hline
\textbf{Average} & \textbf{96.24} & \textbf{76.73} \\
\hline
\end{tabular}
\end{table}

\begin{figure}[t]
  \centering
   \includegraphics[width=0.8\columnwidth]{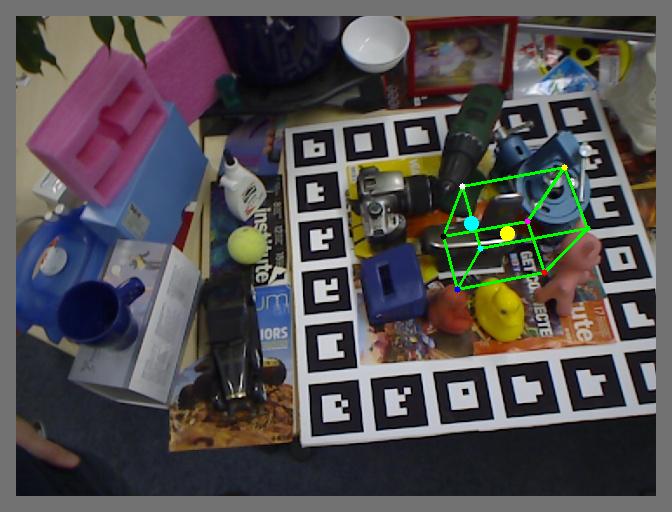}
  \caption{Yellow point represents the projected location of the predicted object center and the cyan point indicates the true object center.}
  \label{fig:depth_error}
 \end{figure}

The substantial drop in accuracy without the keypoint head is primarily attributable to a failure in resolving depth ambiguity. As suggested by Figure \ref{fig:depth_error}, without the explicit spatial anchors provided by the corners, the model lacks the necessary perspective cues to reliably estimate an object's distance. 

\section{Conclusion}
\label{sec:conc}
In this paper, we introduced Yolo-Key-6D, a novel single stage, end-to-end framework for monocular 6D object pose estimation. Our approach builds upon a streamlined YOLO based architecture, eliminating the need for complex multi stage pipelines. We achieve an accuracy of 96.24\% on the standard LINEMOD dataset and 69.41\% on the LINEMOD-Occluded dataset. Our model accomplishes this in real time, running at approximately 63 FPS on a modern GPU. 


\end{document}